\documentclass[a4paper,12pt,onecolumn]{article}

\usepackage{FAS}
\usepackage[T1]{fontenc}
\usepackage[utf8]{inputenc}
\usepackage[english]{babel}
\usepackage{graphicx}
\usepackage{booktabs}
\usepackage{tabularx}
\usepackage{caption}
\usepackage{subcaption}
\usepackage{placeins}
\usepackage{csquotes}
\usepackage{url}
\usepackage{hyperref}

\usepackage[
  backend=biber,
  style=ieee,
  sorting=none,
  doi=true,
  isbn=false,
  date=year,
  maxbibnames=3
]{biblatex}
\AtEveryBibitem{%
  \clearfield{month}%
  \clearfield{day}%
}

\begin{document}
%
\title{Scalable Distributed Simulation-Based Testing for Automated Driving Systems}
%
%
\author{Christian Geller%
\thanks{Christian Geller is doctoral researcher and \textit{ Simulation-Driven Testing} Specialist at the Institute for Automotive
Engineering (ika) at RWTH Aachen University (e-mail: christian.geller@ika.rwth-aachen.de).}\hspace{0.15em},%
\ Benedikt Haas%
\thanks{Benedikt Haas was a student researcher at ika and made major conceptual and technical contributions.}\hspace{0.15em},%
\ and Lutz Eckstein%
\thanks{Lutz Eckstein is professor and director of ika.}
}
\date{}

\maketitle \thispagestyle{empty}

\begin{abstract} Virtual scenario-based testing is a key enabler for validating automated driving systems (ADS) and intelligent transport systems (ITS). However, executing large-scale test suites involving possibly thousands of scenarios remains labor-intensive and difficult to scale. This paper presents an end-to-end, DevOps-driven framework that automates build, deployment, and distributed execution of CARLA-based scenario tests of an ADS on a lightweight Kubernetes cluster. ROS~2 applications are packaged as standardized Kubernetes \texttt{Helm} charts generated from repository specifications, while entire simulation environments are composed declaratively via dynamic \texttt{Helmfile} manifests. The paper describes how a distributed testing workflow can be implemented in Argo Workflows to provision environments, aggregate and batch \mbox{OpenSCENARIO} test cases from configurable sources, execute scenarios in parallel across cluster nodes, and collect logs and resource metrics. In an evaluation on a multi-node K3s cluster running 200 scenarios, the best configuration speeds up end-to-end workflow time by more than a factor of eight compared to a sequential baseline. The results demonstrate significant gains in end-to-end execution time and quantify trade-offs between parallelism, orchestration overhead, and cluster stability. The framework is further demonstrated in a real-world ADS test application with connections to scenario sources and downstream evaluation modules. This demonstrates that the approach provides a strong foundation not only for scalable simulation testing, but also for generating traceable evidence that can support safety arguments.
\end{abstract}


\section{Introduction} 
Automated driving systems (ADS) are high-risk systems and are required to operate reliably under diverse and dynamic real-world conditions, leading to fast-evolving software requirements and regression testing demands~\cite{Miquet25}. With short development cycles and frequent software updates, large test suites must be executed repeatedly across the development lifecycle and, ideally, in parallel at scale~\cite{Nouri25,Kallweit21}. At the same time, test suites evolve continuously as new test cases are added and existing ones are refined. Simulation-based validation complements real-world testing by enabling controlled, repeatable execution of challenging scenarios~\cite{Dona22,Zhong21,Kallweit21}.

\begin{figure}[htb]
    \centering
    \includegraphics[width=0.90\linewidth]{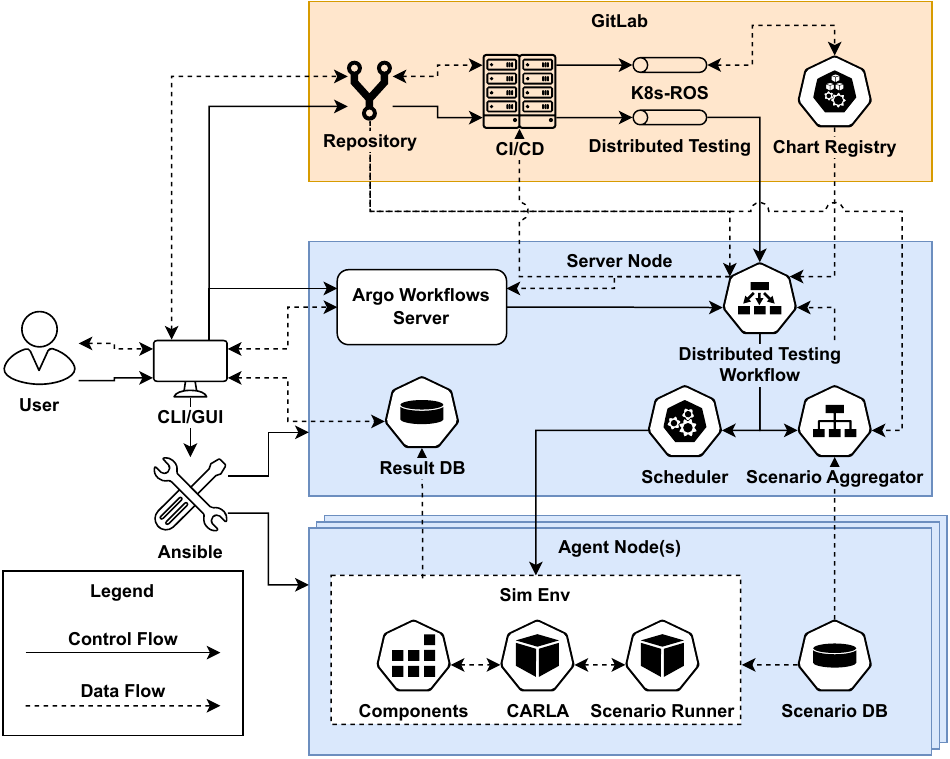}
    \caption{High-level overview of the proposed toolchain and cluster-based execution concept for automated and distributed scenario-based simulation testing.}
    \label{fig:introduction:architecture}
\end{figure}

In practice, the goal is to embed simulation-driven tests into the continuous development process, so that simulation tests are triggered automatically. This enables fast feedback and supports generating reproducible evidence for safety validation~\cite{EC_2022_1426}. However, scaling from single development units to robust, repeatable execution at scale remains challenging due to gaps in automation of the testing environment and test execution workflow~\cite{Casalicchio19,Nalic21}.

This paper proposes an end-to-end, DevOps-driven framework that covers the full chain from software build to automated execution and test result collection. It automates provisioning, execution, and teardown of simulation test environments and distributes scenario execution across a lightweight Kubernetes cluster (Fig.~\ref{fig:introduction:architecture}). ROS~2 applications are integrated via template-based packaging into standardized \texttt{Helm} charts~\cite{Helm25}, while complete simulation environments are composed declaratively via dynamic \texttt{Helmfile} manifests~\cite{Helmfile25}. The evaluation analyzes how orchestration settings influence end-to-end workflow time, overhead, and stability, highlighting practical sweet spots for large-scale regression testing. 

\vspace{0.25cm}
The contributions of this paper are

\begin{itemize}
    \item an architecture for automated and distributed scenario-based simulation testing,
    \item a template design for ROS 2 applications and environments in Kubernetes,
    \item an empirical analysis of end-to-end execution time and cluster resource utilization,
    \item a reference implementation applied to a real ADS testing use case.
\end{itemize}

\newpage
\section{State of the Art}

This section summarizes key foundations and related work in scenario-based simulation testing and DevOps-oriented orchestration approaches. It highlights why existing building blocks such as scenario standards, simulators, and container platforms are not sufficient on their own to achieve end-to-end, large-scale testing for ADS.

Scenario-based validation is widely used to test ADS and their components beyond what can be covered by road or proving ground testing and is also required by current regulations such as (EU) 2022/1426~\cite{EC_2022_1426}. Standards such as OpenSCENARIO and OpenDRIVE enable portable scenario and map definitions, and are commonly combined with software-in-the-loop (SiL) execution to support consistent testing across frequent software revisions and different test environments~\cite{Asam25,Nalic21,Zhong21}. In practice, scenarios are curated from heterogeneous sources, either handcrafted or derived from data-driven recordings, and are stored and versioned in databases like the scenario.center~\cite{scenariocenter24}. For ADS testing, those scenarios must be mapped to specific simulator capabilities and evaluation metrics.

CARLA is frequently used as an open simulation core for ADS research and validation~\cite{Dosovitskiy17}. Building on CARLA, CARLOS~\cite{Geller24} provides an open, modular, and scalable simulation framework that already adopts DevOps principles such as containerized components and automated deployment, and supports multiple use cases including automated testing of ADS software. In this paper, we build on these concepts and extend them toward end-to-end, large-scale, and automated execution of test suites in Kubernetes.

Existing simulation frameworks and scenario execution toolchains provide essential execution primitives, but end-to-end solutions for large-scale regression testing typically address only parts of the overall workflow (e.g., scenario execution, environment setup, or result logging) and still require substantial effort to combine environment composition, scalable execution, and systematic artifact collection~\cite{Casalicchio19,Nalic21}.

To manage the complexity of distributed simulation test stacks, prior work leverages containerization and orchestration concepts. Some approaches implement custom orchestration mechanisms tailored to the application~\cite{Benkendorf25}, while others build on established technologies such as Docker Compose or Kubernetes~\cite{Maller23,Geller24,SYNKROTRON25}. Across these works, distributed execution and modern orchestration improve robustness and scalability, but adapting solutions beyond their target use case often remains effort-intensive~\cite{Casalicchio19}.

\section{Research Approach}
\label{sec:research}

This section motivates the framework design by identifying the practical gap between existing building blocks and the requirements of end-to-end, scalable simulation testing.

Although powerful building blocks including simulators, scenario standards, and container orchestration platforms already exist, there is still a lack of an open, lightweight, and flexible end-to-end approach for scenario-based simulation testing that can be adopted with low effort in custom development processes of ADS. In particular, to the best of our knowledge, there is currently no reference implementation that combines reproducible test environment composition with large-scale scenario execution. Moreover, existing approaches often rely on custom orchestration mechanisms instead of primarily using native Kubernetes primitives such as declarative deployments, scheduling, and workflow-style orchestration.

\newpage
Our approach therefore explicitly considers three aspects. First, the architecture shall support comprehensive test environments, including simulator, system under test, scenario engine, evaluator, and auxiliary services, to be described reproducibly as code while staying modular and extensible. Second, developer integration must be streamlined so that ROS~2 applications can be packaged and deployed consistently across local development and CI with minimal manual effort. Third, scalability shall be addressed comprehensively: large test suites should be executed efficiently in parallel, minimizing orchestration overhead such as repeated startup and teardown, and avoiding overload of cluster management that can delay scheduling or cause instability.

\vspace{0.2cm}
The framework is therefore guided by four design goals:
\begin{enumerate}
    \item[(i)] Environments, workflows, and test selection are declarative and versionable by expressing everything as code.
    \item[(ii)] ROS~2 components follow consistent deployment and runtime patterns through standardized packaging and operational conventions.
    \item[(iii)] Simulation tests run automatically in the ADS development pipeline through CI/CD integration, providing fast feedback and enabling reproducible pre-deployment test campaigns that generate evidence for safety validation.
    \item[(iv)] Distributed scaling is supported through resource-aware parallel execution with explicit mechanisms to observe and control overhead.
\end{enumerate}

Based on these goals, we derive functional and non-functional requirements. Functionally, the framework shall provide complete and isolated simulation environments on demand, assemble scenario suites from configurable sources, and execute standardized scenarios reliably in parallel across cluster nodes. It shall collect and persist structured artifacts such as logs, traces, and resource metrics to support both pass/fail reporting and performance analysis, and it shall tear down environments deterministically even in the presence of failed scenarios or partial workflow retries. Non-functionally, the framework shall ensure reproducibility across software and environment revisions, achieve high throughput without excessive orchestration overhead, and remain usable and maintainable for developers by keeping integration effort low, configuration predictable, and extensions possible without invasive changes to the orchestration core.

\section{Architecture}
\label{sec:architecture}

Figure~\ref{fig:introduction:architecture} summarizes the overall architecture and workflow from source code to CARLA-based distributed scenario execution. Hence, this chapter is structured along four building blocks: (i) standardized packaging of ROS~2 repositories into Kubernetes-ready \texttt{Helm} charts (Sec.~\ref{sec:architecture:k8sros}), (ii) declarative composition of complete simulation environments via dynamically generated \texttt{Helmfile} manifests (Sec.~\ref{sec:architecture:helmfile}), (iii) ADS, scenario and evaluation metric specification based on OpenSCENARIO (Sec.~\ref{sec:architecture:spec}), and (iv) an Argo Workflows-based distributed execution pipeline~\cite{Argo25} for batching, parallel runs, and artifact collection on a lightweight K3s cluster (Sec.~\ref{sec:architecture:workflow}).

\subsection{ROS Applications in Kubernetes}
\label{sec:architecture:k8sros}

ROS~2 projects can be equipped with build pipelines that continuously produce runnable Docker images for development and deployment. In our implementation we rely on the \texttt{docker-ros} toolchain~\cite{dorotos24} to build and publish such container images. The resulting image contains all runtime dependencies and an explicit entrypoint, i.e., the command and arguments required to start the application.

Building on this containerization baseline, we introduce K8s-ROS as an analogous packaging toolchain for Kubernetes. It builds upon a versioned \texttt{Helm} chart template that is reused across repositories and adapted through repository-specific configuration. This template-based approach provides consistent configuration interfaces via \texttt{values} files and enables uniform environment composition (see Sec.~\ref{sec:architecture:helmfile}).

\begin{figure}[htb]
    \centering
    \includegraphics[width=0.9\linewidth]{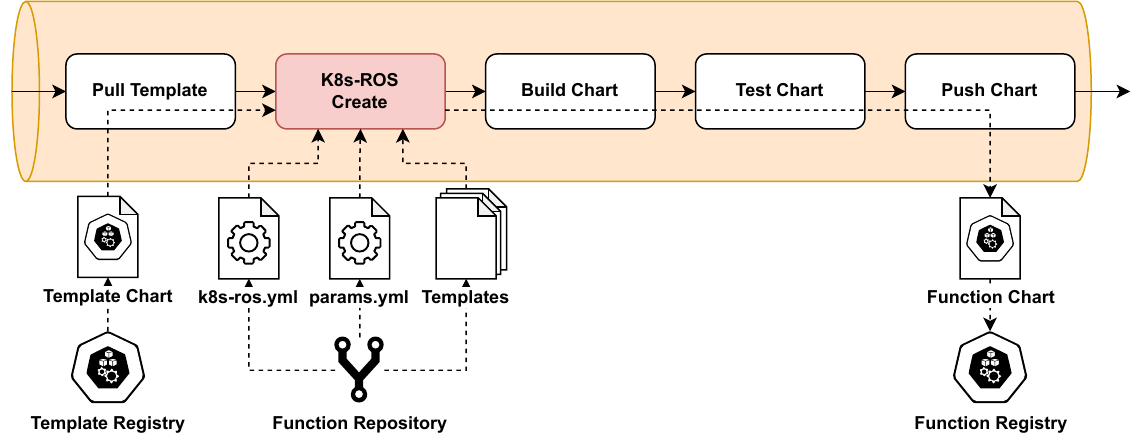}
    \caption{K8s-ROS packaging workflow: CI pipelines customize a versioned \texttt{Helm} chart base-template through configuration and optional overrides; the resulting chart is built, tested, and published to a registry.}
    \label{fig:architecture:k8s-ros}
\end{figure}

Repositories can provide parameter files and other configuration artifacts, as well as optional template overrides to extend or replace parts of the base chart for application-specific needs. In CI, K8s-ROS builds the chart, runs basic validation/tests, and then pushes and versions the resulting artifact in a chart registry. Figure~\ref{fig:architecture:k8s-ros} summarizes the template-based packaging approach to create ready-to-use ROS~2 \texttt{Helm} charts for deployment.

\subsection{Dynamic Simulation Environments}
\label{sec:architecture:helmfile}

A simulation environment consists of multiple components beyond the simulator itself such as the system under test, the scenario runner, and auxiliary services for bridging and observability. The idea of containerized, composable simulation environments is adopted from the CARLOS architecture~\cite{Geller24} and now extended to Kubernetes by packaging ROS~2 components into reusable \texttt{Helm} charts (Sec.~\ref{sec:architecture:k8sros}).

To keep this composition reproducible and manageable, environments are specified using \texttt{Helmfile}. Rather than hardcoding environment variants in scripts, \texttt{Helmfile} allows the environment definition to be parameterized and adjusted from the outside for each test execution. This is essential because moving parts change between runs, including the selected scenario, map, and versions of the ADS.

In our setup, \texttt{Helmfile} is used with a stable base definition and externally supplied values that are merged for a concrete run. The term dynamic \texttt{Helmfile} refers to generating this concrete, run-specific \texttt{Helmfile} configuration from the base definition plus the selected values, so that provisioning and teardown remain reproducible while still being configurable per run.

\subsection{Scenario and Test Specification}
\label{sec:architecture:spec}
Test cases are specified as ASAM OpenX artifacts to keep scenarios portable across simulator cores~\cite{Asam25}. In our implementation, a test case consists of an OpenSCENARIO file and a road representation. Depending on the map, this is either a simulator-native 3D world for known environments or a custom OpenDRIVE file. The OpenSCENARIO file in our setup also includes basic online evaluation metrics and thresholds used for initial pass/fail decisions during scenario execution, while a more detailed a posteriori analysis is performed later on recorded data. In addition, we can reference special runtime configurations or software versions of the system under test. 

To integrate test suites into the workflow, the framework supports multiple scenario sources. Test cases can be stored in versioned repositories, provided through external databases, or retrieved via APIs such as those provided by the scenario.center~\cite{scenariocenter24}. This work demonstrates how such heterogeneous sources can be connected to the execution workflow, while the systematic definition of validity criteria for selecting a test suite remains a separate concern.

For a given test campaign, the framework aggregates the selected test cases from these sources using selectors or API queries, depending on the source. When using scenario.center, the aggregator selects scenarios via a defined GraphQL query, downloads the required OpenX artifacts, and records the selected scenario IDs for reproducible reruns. The resulting scenario list is processed into a common representation that references the OpenSCENARIO file, the map data, and the online evaluation specification. If required, the test run can be executed with an optional ADS configuration override.

During execution, the online evaluation provides the initial pass/fail result for each test case. In parallel, trajectory data is recorded from the simulation and made available through \texttt{omega-prime}, an open-source data model and format for ground-truth traffic data~\cite{OmegaPrime25}. \texttt{omega-prime} builds on ASAM OpenDRIVE and ASAM Open Simulation Interface ground-truth messages and provides a structured access layer for recorded simulation trajectories. These trajectories enable subsequent offline evaluation beyond the initial online checks. Such a posteriori assessment can be complex and is not a direct part of the workflow architecture, but can be attached to the stored trajectory data and is briefly outlined in Sec.~\ref{sec:evaluation:application}. The resulting test case outputs are stored in a central test database to support automated and subsequent analysis.

\subsection{Distributed Testing Workflow}
\label{sec:architecture:workflow}

\begin{figure}[h]
    \centering
    \includegraphics[width=0.9\linewidth]{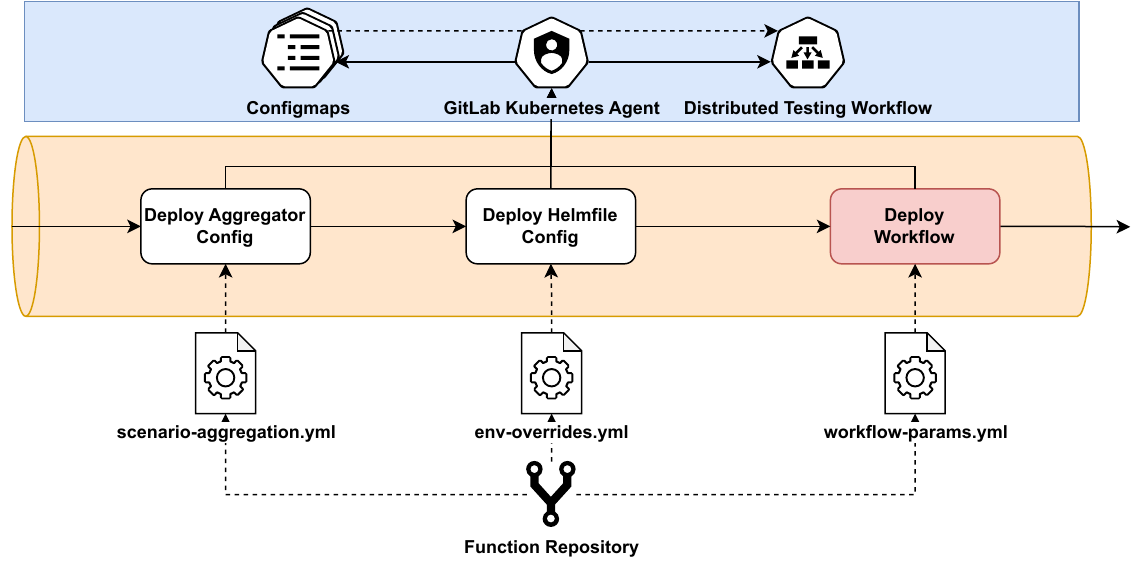}
    \caption{CI-level preparation of a distributed test run: configure scenario aggregation, provide dynamic Helmfile overrides for the test campaign, and submit the Argo workflow.}
    \label{fig:architecture:distributed-testing}
\end{figure}

A distributed test run is started via a three-step pipeline that can be triggered manually or from CI. First, the scenario aggregation configuration described in Sec.~\ref{sec:architecture:spec}~ is deployed to define the list of relevant test cases for the current testing activity. Second, the dynamic \texttt{Helmfile} manifest is configured using potential test-specific value overrides. Third, the pipeline triggers the workflow using execution parameters. Figure~\ref{fig:architecture:distributed-testing} summarizes this preparation and workflow trigger phase.

\begin{figure}[htb]
    \centering
    \includegraphics[width=0.95\linewidth]{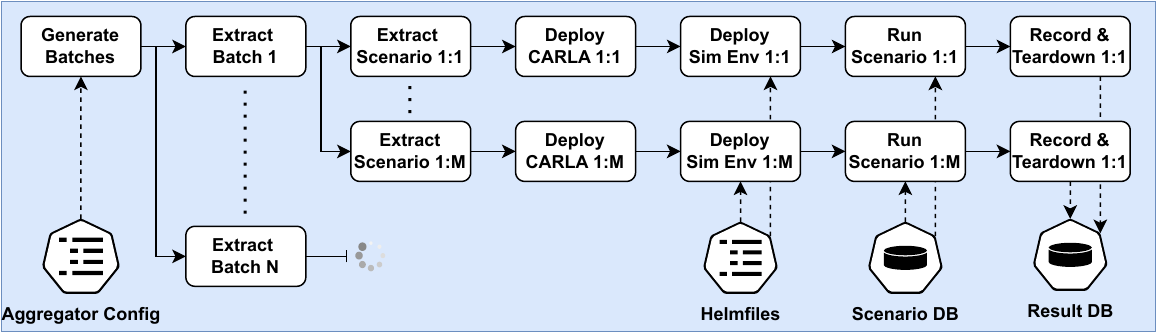}
    \caption{Argo-based execution workflow inside the cluster: generate scenario batches, provision simulator and environment per scenario, run scenarios in parallel, collect artifacts and metrics, and tear down resources.}
    \label{fig:architecture:distributed-testing-workflow}
\end{figure}

The execution itself is orchestrated using Argo Workflows. Figure~\ref{fig:architecture:distributed-testing-workflow}~ shows the concrete workflow executed inside the cluster. The workflow starts by deploying the scenario aggregator, which returns a list of selected test cases, optionally grouped into batches. A batch is a contiguous subset of scenarios that the workflow schedules as one unit. All scenarios of a batch are submitted to the cluster at the same time, but the scheduler starts individual scenarios as resources become available, so scenarios within a batch may still run sequentially. This primarily reduces Kubernetes control-plane pressure by limiting the number of Kubernetes objects created concurrently and by avoiding large spikes of workflow-managed pods. Pods are the Kubernetes execution units, and the control plane manages scheduling. Batches are processed sequentially, while scenarios within a batch can be executed in parallel.

For each scenario execution, the workflow provisions a dedicated simulator instance and deploys the corresponding simulation environment using \texttt{Helmfile}. Each simulator worker pod defines explicit resource requests and limits, including GPU resources where applicable. These resource specifications are used by the Kubernetes scheduler to perform load balancing across Kubernetes nodes: pods are placed on nodes with available capacity; if resources are unavailable, the test case remains pending. Parallelism is controlled by the workflow parameters batch size and the number of simulator instances per node.

During execution, the workflow collects artifacts such as logs, traces, and resource metrics, and persists them for reporting and later analysis. Separating simulation environment lifecycle management from the actual scenario execution makes orchestration overhead measurable and tunable.

\section{Evaluation}
\label{sec:evaluation}

This section first evaluates the proposed framework in a controlled cluster setup and discusses how key orchestration parameters affect end-to-end runtime, utilization, and practical stability. It then illustrates the application of the approach in a more realistic ADS testing setup with heterogeneous scenario sources and subsequent result analysis.

\subsection{Experimental Setup and Metrics}
We evaluate the framework using a controlled experiment series with the CARLA simulator, a full ADS software stack, and 200 OpenSCENARIO test cases. The scenario set is designed to provide heterogeneous execution conditions across different maps and scenario complexities, rather than to represent a complete safety validation suite. This makes the experiments suitable for analyzing orchestration behavior, resource utilization, and workflow scalability. The software stack contains real-time coupled components such as an optimization-based planning module and therefore aims to execute scenarios close to real time.

Experiments are performed on a K3s cluster with three physical compute nodes, including one control-plane/server node and two agent nodes. All three nodes provide 64\,GB of RAM and GPUs with 24\,GB of VRAM, which in principle allows running up to three CARLA simulator instances per node in parallel. Each simulation run deploys the CARLA simulator alongside the full ADS software stack, but in practice the simulator is the primary bottleneck because CARLA is substantially more GPU-intensive than the ADS software. The two key orchestration parameters are the maximum number of simulator instances per node (\texttt{s}) and the workflow batch size (\texttt{b}). We denote a configuration as \texttt{s$X$-b$Y$}, where $X$ is the chosen value for \texttt{s} and $Y$ is the chosen value for \texttt{b}.

We evaluate the end-to-end workflow time from submission to completion, the resulting throughput per hour, and resource utilization for CPU, GPU and RAM as indicators for efficiency. In addition, we qualitatively track stability symptoms under high parallelism, for example failed pods, workflow retries, and reduced control-plane responsiveness.

\subsection{Experiments and Results}
\label{sec:evaluation:experiments}

We compare all distributed configurations against a sequential baseline (\texttt{s1-b1}). Table~\ref{tab:workflow-times} provides an overview of end-to-end workflow time, throughput, and average resource utilization across all tested configurations.

\paragraph{Effect of simulator instances per node:}
Increasing the number of simulator instances per node reduces workflow time drastically by increasing parallelism. With a fixed batch size of 9, the end-to-end time decreases from 136\,min (\texttt{s1-b9}) to 95.0\,min (\texttt{s2-b9}) to 58.7\,min (\texttt{s3-b9}). Notably, the speed-up is not strictly proportional to the increase in \texttt{s}, indicating non-negligible overheads and resource contention effects.

However, running three simulators per node with graphical output pushes compute resources close to saturation: the average GPU utilization rises from 22.46\% (\texttt{s2-b9}) to 37.81\% (\texttt{s3-b9}), with peaks reaching 100\%. In addition, the scenario runner repeatedly reports execution delays, indicating that the simulation cannot keep real time and that higher throughput comes at the cost of a reduced real-time factor.

\vspace{0.75cm}

\begin{table}[h]
\centering
\small
\setlength{\tabcolsep}{4pt}
\begin{tabularx}{\linewidth}{l*{6}{>{\centering\arraybackslash}X}}
\toprule
\textbf{Metric} & \textbf{\texttt{s1-b1}} & \textbf{\texttt{s1-b9}} & \textbf{\texttt{s2-b9}} & \textbf{\texttt{s3-b9}} & \textbf{\texttt{s2-b60}} & \textbf{\texttt{s2-b200}} \\
\midrule
Workflow time [$\mathrm{min}$] & 475.0 & 136.0 & 95.0 & 58.7 & 67.7 & 65.2 \\
Throughput [$h^{-1}$]     &  25.3 &  88.3 & 126.3 & 204.5 & 177.3 & 184.0 \\
CPU utilization [\%]     & 11.72 & 16.63 & 30.09 & 37.45 & 45.29 & 42.47 \\
RAM utilization [\%]     &  8.99 &  9.21 & 14.28 & 18.40 & 27.68 & 22.15 \\
GPU utilization [\%]     & 11.73 & 18.75 & 22.46 & 37.81 & 22.39 & 22.39 \\
\bottomrule
\end{tabularx}
\caption{Aggregated metrics for the experiments executing 200 scenarios per workflow. We denote a workflow configuration as \texttt{s$X$-b$Y$} where $X$ is the number of simulator instances per node and $Y$ is the workflow batch size.}
\label{tab:workflow-times}
\end{table}

\paragraph{Effect of batch size:}
Batch size controls how scenarios are allocated and submitted to the cluster, and larger batches generally reduce scheduling and startup overheads. The sequential configuration \texttt{s1-b1} illustrates that overhead can dominate at very small batch sizes: executing 200 scenarios requires 475\,min (\,$\approx 8$\,h), i.e., about 142\,s per scenario on average, while scenario execution in this controlled experiment setup is limited by a 35\,s timeout. This is reflected by long idle intervals in the resource traces, where CPU and GPU utilization repeatedly drops to nearly zero between scenario runs.

With a fixed number of two simulators per node, increasing batch size from 9 (\texttt{s2-b9}) to 60 (\texttt{s2-b60}) reduces end-to-end time from 95.0\,min to 67.7\,min; correspondingly, idle phases become shorter and GPU utilization becomes more continuous. At the same time, extremely large batches can overload the system. With batch size 200 (\texttt{s2-b200}), the runtime improves only marginally to 65.2\,min, but control-plane responsiveness degrades and
the framework becomes unstable, consistent with control-plane pressure caused by large
numbers of workflow-managed pods.

\begin{figure}[htb]
  \centering
  \begin{subfigure}[t]{0.3\linewidth}
    \centering
    \includegraphics[width=\linewidth]{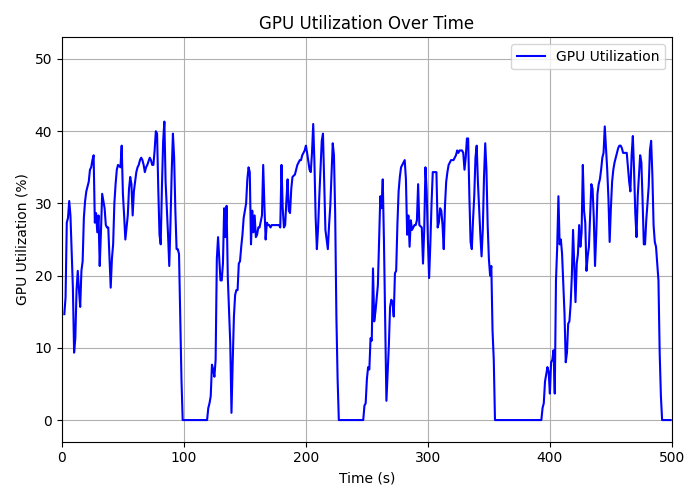}
    \caption{\texttt{s1-b9}}
  \end{subfigure}
  \begin{subfigure}[t]{0.3\linewidth}
    \centering
    \includegraphics[width=\linewidth]{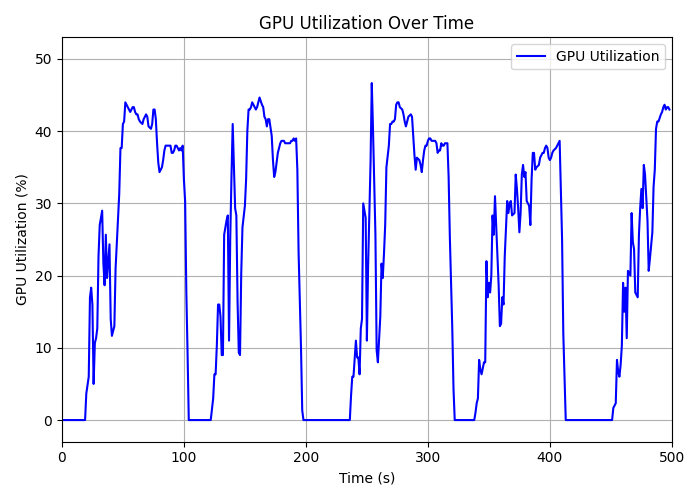}
    \caption{\texttt{s2-b9}}
  \end{subfigure}
  \begin{subfigure}[t]{0.3\linewidth}
    \centering
    \includegraphics[width=\linewidth]{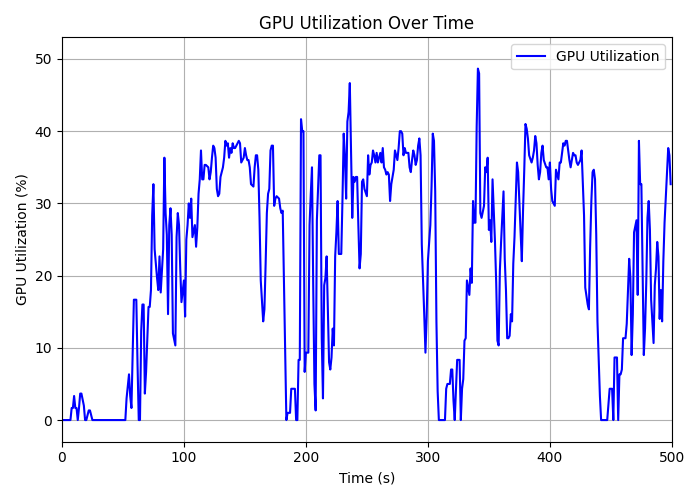}
    \caption{\texttt{s2-b60}}
  \end{subfigure}  
  \caption{GPU utilization over time for three configurations, illustrating reduced idle phases with larger batch sizes.}
  \label{fig:eval:gpu-comparison}
\end{figure}

To illustrate these effects, we compare GPU utilization traces in Fig.~\ref{fig:eval:gpu-comparison}. The plots highlight how settings affect idle phases and how continuously the GPUs are utilized during the workflow.

The traces indicate two primary mechanisms for increasing GPU utilization: larger batches (\texttt{b}) reduce idle gaps, whereas higher per-node parallelism (\texttt{s}) increases concurrent simulator execution but only modestly raises average GPU utilization in our experiments.

These findings suggest treating \texttt{s} and \texttt{b} as tuning parameters. Increase \texttt{s} to improve throughput while keeping real-time execution within acceptable bounds for the intended test fidelity, which is critical because parts of the ADS are real-time coupled. Increase \texttt{b} to reduce orchestration overhead while maintaining stable control-plane responsiveness. In our experiments, medium batch sizes provide most of the overhead reduction, whereas very large batches show diminishing returns and can compromise stability. 

\newpage

In summary, the key observations are:

\begin{itemize}
    \item \textbf{Distributed execution improves throughput substantially:} compared to the sequential, single-simulator configuration \texttt{s1-b1}, all distributed configurations reduce end-to-end time and increase throughput by a factor of 3--8.
    \item \textbf{Increasing simulator instances} per node (\texttt{s}) \textbf{improves throughput} but can \textbf{reduce real-time factor}: at higher \texttt{s}, resources approach saturation and real-time behavior can no longer be guaranteed.
    \item \textbf{Increasing batch size} (\texttt{b}) reduces orchestration \textbf{overhead}: larger batches reduce idle time, but can also \textbf{affect cluster stability} due to control-plane pressure.
\end{itemize}

\subsection{Application to ADS Testing}
\label{sec:evaluation:application}

Building on the experiments described in Sec.~\ref{sec:evaluation:experiments}, we also apply the framework to a more realistic ADS testing setup. This application serves as an illustrative use case, not as a complete safety argument or final validation campaign. It shows how an actual research ADS and heterogeneous scenario sources can be integrated into the proposed architecture, while also demonstrating how recorded trajectories and test metadata support subsequent result analysis. The exact selection of scenarios, metrics, and acceptance criteria is therefore not the main contribution of this paper, but demonstrates the type of end-to-end testing process that the framework is designed to support.

The system under test is our ROS~2-based automated driving software stack that is also used in ongoing real-world operation of our research vehicle karl.~\cite{karl26}. The stack consists of several functional modules, including perception, localization, planning, control, and supporting infrastructure components. Before deployment setup, all ADS components go through the packaging and composition steps described in Secs.~\ref{sec:architecture:k8sros} and~\ref{sec:architecture:helmfile}: each component is containerized, packaged as an individual \texttt{Helm} chart, and integrated into a common stack-level \texttt{Helmfile}. Consequently, each scenario test run later deploys a dedicated pod for every ADS component, together with the simulator interfaces, scenario runner, evaluator, and auxiliary services. This mirrors the intended operational structure of the framework, where the simulated test environment is not a manually assembled prototype, but a reproducible deployment artifact that can be versioned and triggered from a CI/CD process as well.

\begin{figure}[htb]
    \centering
    \includegraphics[width=0.85\linewidth]{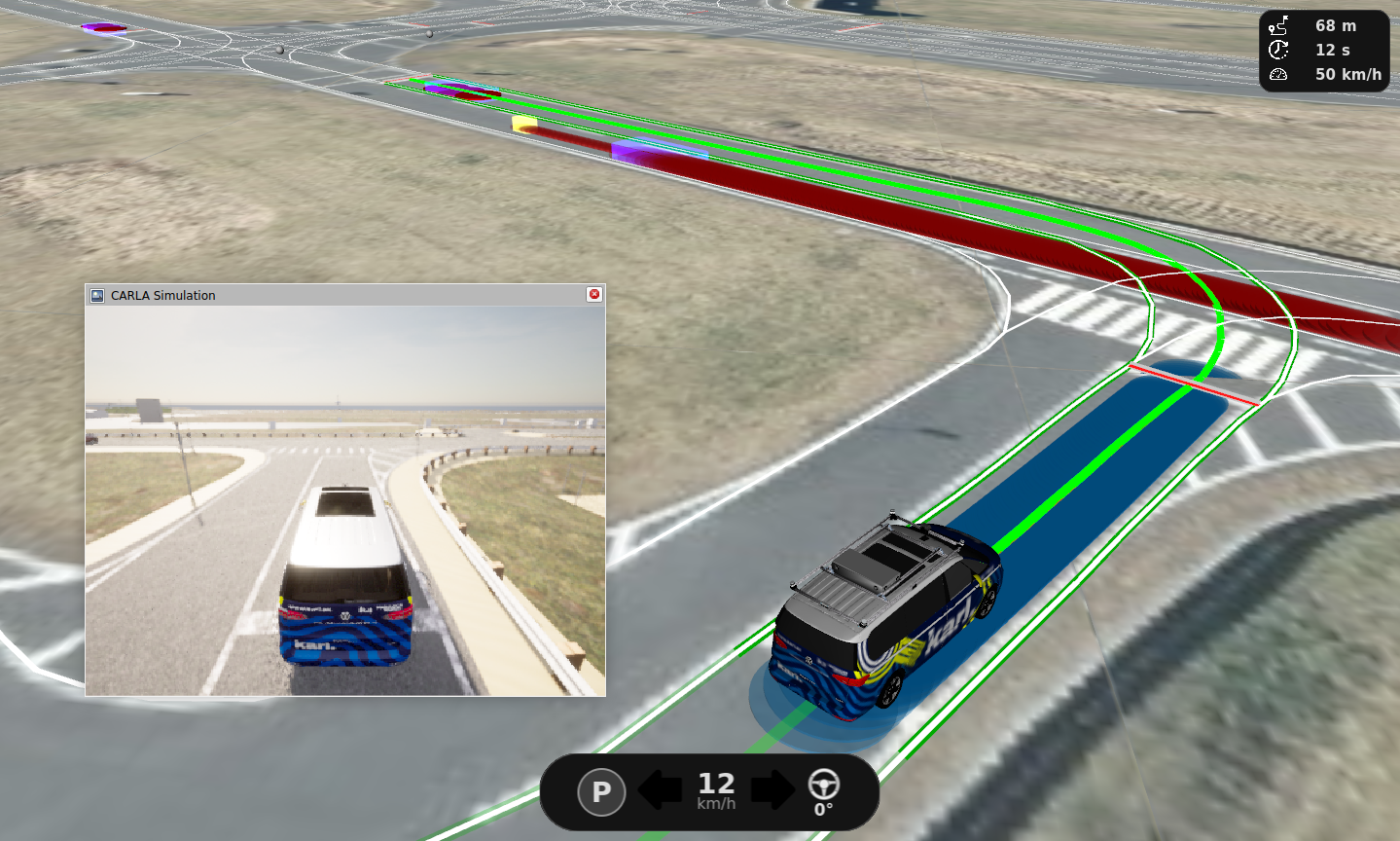}
    \caption{Exemplary left-turn scenario at the Aldenhoven Testing Center with the karl. research vehicle in CARLA, used to test our ROS~2-based automated driving software stack. The simulation is deployed as part of an Argo Workflows-based execution in a Kubernetes cluster.}
    \label{fig:application:left-turn-carla}
\end{figure}

To illustrate how this reproducible deployment appears in a concrete simulation run, Fig.~\ref{fig:application:left-turn-carla} shows an exemplary instance for the described application context.

In contrast to the preceding experiments, which use generated scenarios mainly to stress the execution workflow, this application uses a more meaningful scenario database for a concrete operational design domain. The purpose is to demonstrate that scenarios can be aggregated from heterogeneous sources within the same execution pipeline. Data-driven scenarios are derived from real-world traffic observations and can, among other sources, be provided via scenario.center~\cite{scenariocenter24}. In our example, these data sources include real-world recordings from the karl. research vehicle as well as the inD drone dataset~\cite{inD20}. Relevant scenarios can be selected using GraphQL queries to scenario.center, either directly at workflow runtime or a priori to create a fixed and reproducible test database. This mechanism is complemented by manually designed scenarios generated using \texttt{simple-scenario}~\cite{simplescenario25}, which encode expert knowledge about relevant situations in the intended operational design domain and cover cases that may not be sufficiently represented in recorded data.

For the presented application, this process results in an a priori scenario database containing 81 concrete test scenarios. Although the number is smaller than in the preceding experiment series, these scenarios are separate and selected more deliberately for the considered ADS testing context. The set is not intended to be complete or optimal from a safety validation perspective, but provides a useful test suite for iterative development.

The Kubernetes workflow configuration for this application is chosen based on the findings of the experiments in Sec.~\ref{sec:evaluation:experiments}, but scaled to an even larger cluster. The setup uses five Kubernetes worker nodes and configures up to two simulator instances per node with a batch size of $16$, following the trade-off identified above between high utilization and control-plane stability. Automatic trajectory data recording is enabled as described in Sec.~\ref{sec:architecture:spec}, so that each simulation run produces data within the \texttt{omega-prime} format for subsequent offline evaluation. The overall campaign completes the 81 scenarios in approximately $2 385\,\mathrm{s}$ of wall-clock time, corresponding to a throughput of about $2.0$ scenarios per minute, with an average simulated scenario duration of $56.7\,\mathrm{s}$.

The recorded output data can also support subsequent a posteriori evaluation. In this paper, we include such an analysis as an illustrative extension of the workflow, while keeping the focus on the execution architecture. During evaluation, test-run-specific metadata are collected that describe the ADS, the scenario, and the simulation as the three core elements of the test execution. Established scenario analysis tools used within scenario.center can further enrich these metadata with information about dynamic object interactions. Based on these metadata and the previously stored \texttt{omega-prime} trajectory data, an additional evaluation module adaptively selects situation-dependent metrics and thresholds instead of applying one global criterion set to all test cases.

Following a concept from scenario.center, each executed test scenario is first decomposed into base scenarios, each representing a more interpretable dynamic interaction. The selected metrics and criteria are then applied per base scenario, yielding more situation-specific pass/fail decisions.

\begin{figure}[htb]
    \centering
    \includegraphics[width=0.90\linewidth]{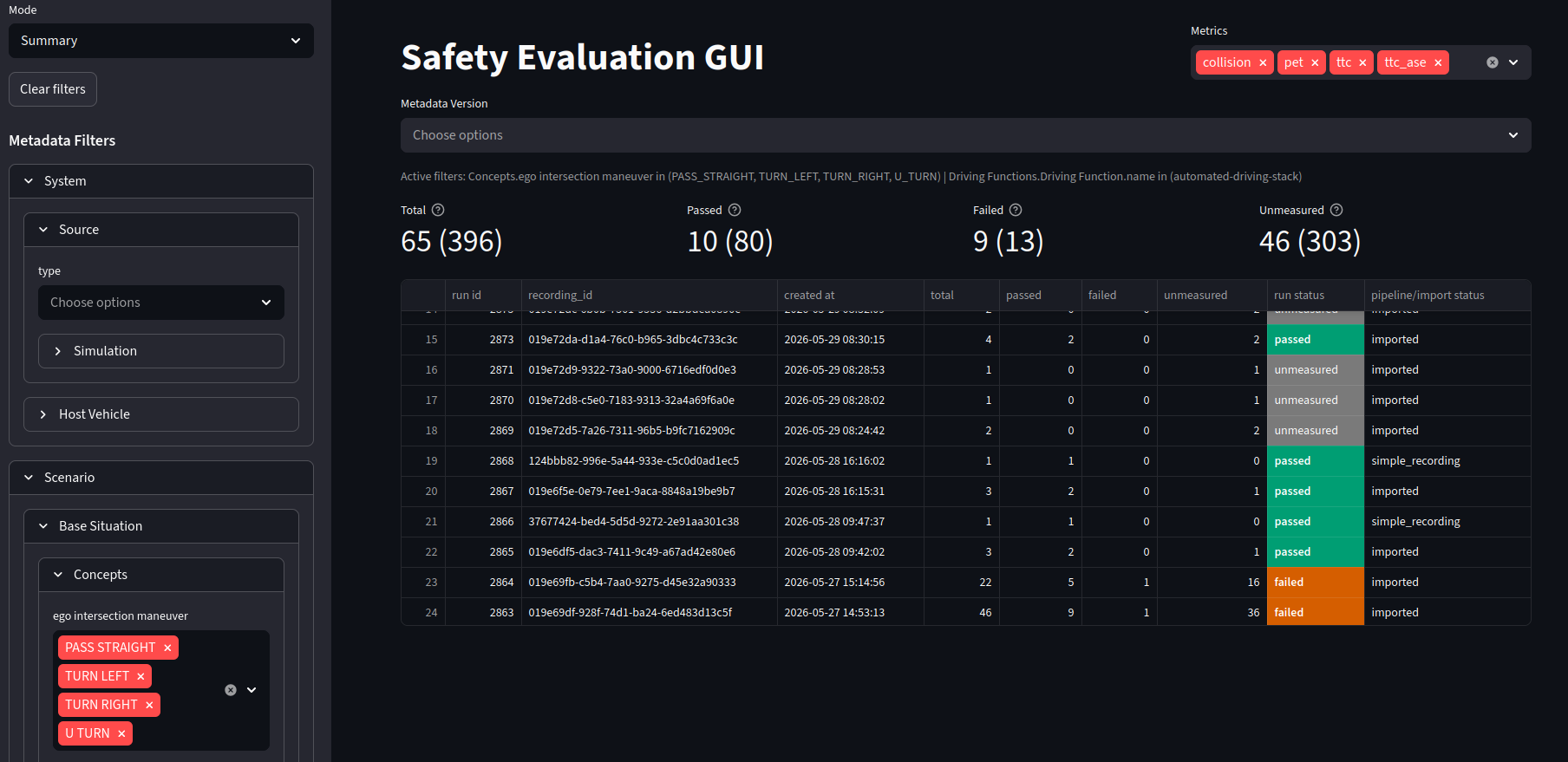}
    \caption{Example visualization of subsequent safety evaluation results for executed test runs. The view shows the number of passed and failed checks on the decomposed base-scenario level and supports flexible analysis through selectable metrics and metadata filters.}
    \label{fig:eval:safety-gui}
\end{figure}

All resulting evaluation data, including test-run metadata, base scenario decomposition, metric values, criteria, and evaluation outcomes, are stored in a PostgreSQL result database. This database can be queried for further processing and analysis, or the results can be explored through a safety evaluation GUI shown in Fig.~\ref{fig:eval:safety-gui}.

\vspace{0.2cm}

In the shown example, 81 test cases were executed initially. After the live evaluation and initial consistency checks, 65 test runs were forwarded to the subsequent a posteriori evaluation, where they were decomposed into 396 base scenarios before being evaluated by up to four metric checks. The safety evaluation GUI supports three main views: an overview, as shown in Fig.~\ref{fig:eval:safety-gui}, which summarizes the number of passed and failed checks; a coverage view, which shows how many tests are available for individual metadata categories and how successful they are within each category; and a developer view, which provides a detailed view on individual base scenarios, including debugging visualizations. In addition, the GUI enables comparisons between different metadata sets, for example real-world and simulated scenarios, different vehicle setups, or different software versions. The focus of this example is not on the specific numerical results, but on illustrating how additional evaluation pipelines can be integrated into the distributed testing workflow.

\section{Conclusion}
\label{sec:conclusion}
This paper presented an end-to-end DevOps-driven framework for distributed scenario-based simulation testing of ADS on a lightweight Kubernetes cluster. The approach standardizes ROS~2 application packaging into Helm charts (K8s-ROS), composes simulation environments via dynamic \texttt{Helmfile} manifests, and orchestrates large scenario suites with Argo Workflows for parallel execution and artifact collection.

The experimental results demonstrate that the framework is practically applicable: a large number of test cases can be executed robustly on a multi-node K3s cluster, and distributed execution substantially reduces end-to-end workflow time compared to a sequential setup while improving overall resource utilization. The real-world application further illustrates how the same workflow can be connected to heterogeneous scenario sources and subsequent evaluation pipelines, including metadata-based analysis and scenario-level safety assessment.

Future work will strengthen monitoring and observability using Prometheus-based metrics and extend resource modeling from the simulator to all ADS modules, so that scheduling can be improved. In addition, the briefly described evaluation framework could be integrated more tightly into the cluster, for example as an additional workflow step connected to a cluster-managed result database. This would further support traceability from simulation execution to evaluation results and provide a basis for larger data-driven regression test campaigns in ADS development.

\printbibliography

\end{document}